\documentclass{ifacconf}

\usepackage{graphicx}      
\usepackage{natbib}        

\usepackage{amsmath}
\usepackage{hhline}
\usepackage{multirow}
\usepackage{eucal}
\usepackage{amssymb}
\usepackage{enumerate}
\usepackage{subfigure}
\usepackage{color}
\usepackage{bm}

\newcommand*{\QEDA}{\hfill\ensuremath{\blacksquare}}%

\newcommand{\q}{\mathbf{q}}
\newcommand{\bq}{\bar{\mathbf{q}}}
\newcommand{\T}{\boldsymbol{\tau}}
\newcommand{\bT}{\bm{T}}
\newcommand{\B}{{\mathbf{B}}}
\newcommand{\M}{\mathbf{M}}
\newcommand{\C}{\mathbf{C}}
\newcommand{\g}{\mathbf{g}}
\newcommand{\bM}{\bar{\mathbf{M}}}
\newcommand{\bC}{\bar{\mathbf{C}}}
\newcommand{\bg}{\bar{\mathbf{g}}}
\newcommand{\bphi}{\bm{\phi}}
\newcommand{\y}{\mathbf{y}}

\newcommand{\uu}{\mathbf{u}}
\newcommand{\bzero}{\mathbf{0}}
\newcommand{\K}{\mathbf{K}}
\newcommand{\D}{\mathbf{D}}

\newcommand{\x}{\mathbf{x}}
\newcommand{\bI}{\mathbf{I}}

\begin{document}
\begin{frontmatter}

\title{Oscillation Damping Control  of Pendulum-like  Manipulation Platform using Moving Masses\thanksref{footnoteinfo}} 

\thanks[footnoteinfo]{The funding of the European Commission to the AEROARMS project under the H2020 Programme (Grant Agreement 644271) is acknowledged.}

\author[First]{Min Jun Kim} 
\author[First,Second,Third]{Jianjie Lin} 
\author[First]{Konstantin Kondak}
\author[First,Second]{Dongheui Lee}
\author[First]{Christian Ott}

\address[First]{Institute of Robotics and Mechatronics, German Aerospace Center (DLR), Wessling, Germany (e-mail: minjun.kim@dlr.dr).}
\address[Second]{Chair of Automatic Control Engineering, Technical University of Munich (TUM), Munich, Germany}
\address[Third]{Fortiss Institute, Munich, Germany}

\begin{abstract}                
	This paper presents an approach to damp out the oscillatory motion of the pendulum-like hanging platform on which a robotic manipulator is mounted. To this end, moving masses were installed on top of the platform. In this paper, asymptotic stability of the platform (which implies oscillation damping) is achieved by designing reference acceleration of the moving masses properly. A main feature of this work is that we can achieve asymptotic stability of not only the platform, but also the moving masses, which may be challenging due to the under-actuation nature. The proposed scheme is validated by the simulation studies.
\end{abstract}

\begin{keyword}
Under-actuated system, asymptotic stability, partial feedback linearization
\end{keyword}

\end{frontmatter}

\section{Introduction}

\begin{figure}
	\centering
	\subfigure[Heli-manipulator system developed in DLR]{
	\centering
	\includegraphics[scale=0.70]{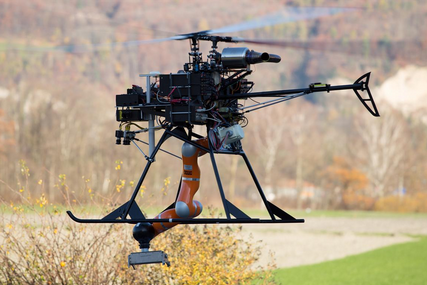}	} \\
	\subfigure[]{
	\centering	
	\includegraphics[scale=0.55]{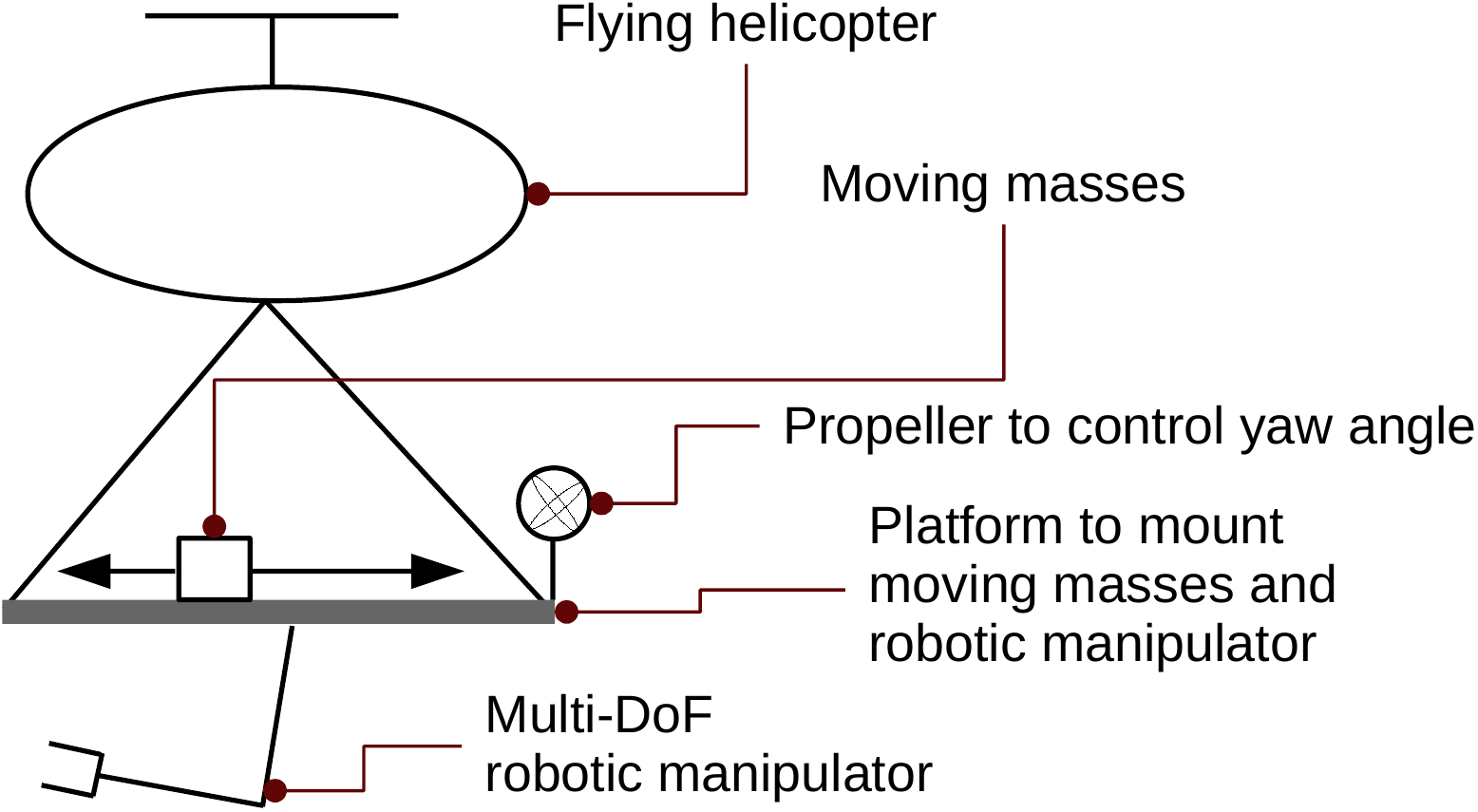}	}
	\caption{Schematic diagram of the system of interest.}
		\label{fig:configuration}
\end{figure}

Control of robotic manipulators on a fixed or wheeled base is well studied in the robot control community; see, e.g., \cite{borst2009rollin, englsberger2014overview, kim2015bringing, kim2017enhancing}. In order to further increase the workspace of the manipulation, a recently proposed solution is given by the combination of a robotic manipulator with a flying base (Fig. \ref{fig:configuration}a), e.g. a helicopter \cite{kondak2013closed, huber2013first, kim2018stabilizing}. However it has a drawback due to the large rotor blades which may cause safety issues when operating in a complex environment. To alleviate this drawback, a new configuration, which can decouple the helicopter and manipulator, is proposed. In this configuration, the robotic manipulator is mounted on a floating platform, which can be suspended on the helicopter by means of wires (Fig. \ref{fig:configuration}b). If the flying helicopter is controlled to be fixed, then the hanging platform is nothing but the oscillating pendulum.

In the control point of view, the main challenge is the oscillations caused by the external force and the movement of the robotic manipulator. \cite{potter2015planar, chen2016swing, vyhlidal2017time} solved similar problems by controlling the base. In this paper, because the precise control of the base (helicopter) is not so trivial for our system, additional moving masses are installed to damp out the oscillations of the pendulum-like hanging platform. Furthermore, to reduce the complexity of the problem, so that we can focus only on the oscillation damping of the platform, propellers are considered to control yaw angle of the platform. Notice that the proposed configuration is an under-actuated mechanical system with a fewer control input than the degrees of freedom (DoF). 

Control of under-actuated systems is still an active research field, as the stabilization is not always trivial \cite{shiriaev2014controlled, lee2015robust}. \cite{spong1997control} introduced the energy-based methods with saturation function to swing up the Acrobot, which is a two-link planar robot with one actuator at the elbow. 
\cite{albu2012energy} have also applied the potential energy shaping control for the underactuated Euler-Lagrange systems by introducing a new feedback state variable, which are statically equivalent to the non-collocated state variables. However from the configuration (Fig. \ref{fig:configuration}b), it can be seen that when the system is stable at a fixed height, the potential energy reaches a constant value, no matter where are the moving masses at the platform. Therefore, the moving masses can have multiple equilibrium points, which is not appealing.  Based on this observation, it can be concluded that (potential) energy shaping method is not suitable for this configuration.
\cite{spong1994partial} made use of the collocated/non-collocated partial feedback linearization (PFL) to control the output dynamics by exploiting the strong inertial coupling under a certain condition. However the remaining dynamic after PFL (internal dynamics) should be carefully taken into account to achieve asymptotic stability of the overall dynamics.


In this paper, we present an extension of PFL technique to asymptotically stabilize the output dynamics as well as internal dynamics. In this paper, within the PFL framework, the dynamics of the moving mass is the output dynamics, and that of the platform orientation is the internal dynamics which does not have collocated actuation. To overcome the absence of the actuation, the internal dynamics is accessed using the reference acceleration of the moving masses. By designing it properly, the internal dynamics as well as the output dynamics can be asymptotically stabilized at the same time.  The proposed approach is verified through simulation studies.

The rest the paper is organized as follows. Section \ref{sec:moelding_problem} introduces the modeling of  the system and states the problem definition. Moreover, a motivating example that shows limitation of the standard PFL approach will be shown. To overcome the limitation, the extension of PFL with stabilization of internal dynamics will be proposed in Section \ref{sec:control}. Simulation validation will be illustrated in Section \ref{sec:simulation}. Section \ref{sec:conclusion} concludes the paper.

\section{System Modeling and Problem Definition }
\label{sec:moelding_problem}

\subsection{System description and control goal}

A schematic diagram of the system of interest is shown in Fig. \ref{fig:configuration}b. A robotic manipulator is mounted on the platform which is suspended by means of the wire which can be connected to, for example, a flying helicopter or a crane.  Because the pendulum-like platform may oscillate due to the external disturbances (e.g. wind gust) and the movement of the robotic manipulator, the moving masses are installed to damp out the oscillation. The main control goal is to achieve oscillation damping of the platform (i.e., achieving zero roll and pitch angles) using moving masses. Propellers are considered to control yaw angle independently (see Fig. \ref{fig:configuration}b), so that we can focus only on the oscillation damping problem. 

The equation of motion can be expressed as
\begin{align}
\M(\q)\ddot{\q}+\C(\q,\dot{\q})\dot{\q}+\g(\q)=\T
\label{eq:whole1}
\end{align}
where the vector {$\q=[\q_p^T \; \q_m^T \; \q_r^T]^T \in \mathbb{R}^{3+2+n}$} consists of platform orientation $\q_p \in \mathbb{R}^3$,  moving mass coordinates $\q_m \in \mathbb{R}^2$, and the robotic arm coordinates $\q_r \in \mathbb{R}^n$. In this paper, $\q_p$ is composed of roll, pitch, and yaw angles $\alpha$, $\beta$, $\gamma$; i.e., $\q_p=[\alpha \; \beta \; \gamma]^T$.
The torque $\T$ is given by
\begin{align}
\begin{matrix}
\T=\begin{pmatrix}
\mathbf{0}_{2 \times 1} \\ \mathbf{u}
\end{pmatrix}, & \text{with\:\:} \mathbf{u}=\begin{pmatrix}
\tau_{yaw} \\ \T_m \\ \T_r
\end{pmatrix}, \T_m \in \mathbb{R}^2 
\text{\; and \;} \T_r \in \mathbb{R}^n
\end{matrix}.
\end{align}
Equivalently, $\T$ can be represented as
\begin{align}
\T = \B \uu,
\end{align}
where 
\begin{align}
\label{eq:def_B}
\B = \left[
\begin{array}{c}
\bzero_{2 \times (3+n)} \\
 \bm{I}_{3+n}
\end{array}
\right] \in \Re^{(6+n)\times(3+n)}.
\end{align}
$\tau_{yaw}$ is the torque generated by propellers to control yaw angle. $\T_m$ and $\T_r$ are forces/torques to control moving masses and robotic manipulator, respectively. Note that there is no actuation along the roll ($\alpha$) and pitch ($\beta$) angles of the platform, and this under-actuated nature makes it difficult to achieve the control goal which can be summarized as follows.
\begin{enumerate}
\item {Robotic manipulator variable $\q_r$ converges to the desired value $\q_r^{des}$.}
\item {Platform yaw angle $\gamma$ converges to the desired value $\gamma^{des}$.}
\item {Platform roll and pitch angles $\alpha$, $\beta$ converge to zero (oscillation damping and gravity compensation of the pendulum-like platform).}
\item {Moving mass variable $\q_m$ converges to a certain equilibrium point $\q_m^{*}$ at which the moving masses balance the static gravity torque when $\q_r=\q_r^{des}$. Hereinafter, $\q_m^*$ will be assumed to be known as it can be pre-calculated.}
\end{enumerate}

\subsection{Motivating example using standard PFL approach}

In this section, a standard PFL technique will be applied to achieve oscillation damping of the platform (i.e., $\alpha=\beta=0$). To begin with, let us define the following notations for later convenience.
\begin{itemize}
\item {$\bphi=[\alpha \; \beta]^T$: Roll and pitch angles.}
\item {$\M_{\phi\phi}$: First 2 by 2 block matrix of $\M$.}
\item {$\M_{\phi m}$: First 2 by (4:5) block matrix of $\M$. Maps $\ddot{\q}_m$ to $\ddot{\bphi}$.}
\item {$\C_{\phi \phi} \in \Re^{2 \times 2}$ and $\C_{\phi m} \in \Re^{2 \times 2}$ are defined similarly.}
\item {$\g_{\phi}$: First two components of $\g$.}
\end{itemize}

To apply PFL, define the output variable by
\begin{align}
\label{eq:y_PFL}
\y = 
\left(
\begin{array}{c}
\gamma \\
\q_m \\
\q_r
\end{array}
\right)
=
\B^T \q 
\end{align}
with the selection matrix $\B$ defined in (\ref{eq:def_B}). Then, 
\begin{align}
\nonumber \ddot{\y} =& \B^{T} \ddot{\q} \\
              =& \B^{T} ( \M^{-1}  -\C \dot{\q} - \g  + \B \uu ).
\end{align}
Using
\begin{align}
\uu = (\B^{T} \M^{-1} \B)^{-1}  \left( \B^T  \M^{-1}  (\C \dot{\q} + \g) + \ddot{\y}^{ref} \right),
\end{align}
where $\ddot{\y} = [\ddot{\gamma}^{ref} \; \ddot{\q}_m^{ref,T} \; \ddot{\q}_r^{ref,T}]^T$, we obtain
\begin{align}
\ddot{\y} = \ddot{\y}^{ref}.
\end{align}
Defining
\begin{align}
\label{eq:PFL_gamma_ref}
\ddot{\gamma}^{ref} = -D_\gamma \dot{\gamma} - K_\gamma (\gamma - \gamma^{des}),\\
\label{eq:PFL_qr_ref}
\ddot{\q}_r^{ref} =  -\D_r \dot{\q}_r - \K_r (\q_r - \q_r^{des}),
\end{align}
the configuration of the robotic manipulator and yaw angle will converge to the desired values. $\ddot{\q}_m^{ref}$ will be designed shortly to damp out the oscillation of the platform.

At this point, we introduce a well-known result from the cascade control literature \cite{seibert1990global}.
\begin{thm}
	\label{thm:cascade}
	Consider a system
	\begin{align}
	\label{eq:cascade_1}
	\dot{\x}_1 =& \bm{f}_1(\x_1), \\
	\label{eq:cascade_2}
	\dot{\x}_2 =& \bm{f}_2(\x_1,\x_2) .
	\end{align}
	If $\dot{\x}_1=\bm{f}_1(\x_1)$ is locally asymptotically stable to $\x_1=\bzero$ and $\dot{\x}_2=\bm{f}_2(\bzero,\x_2)$ is locally asymptotically stable to $\x_2=\bzero$, then (\ref{eq:cascade_1})-(\ref{eq:cascade_2}) is locally asymptotically stable to $\x_1=\bzero$ and $\x_2=\bzero$.
\end{thm}

By the cascade control theory (Theorem \ref{thm:cascade}), $\gamma \equiv  \gamma^{d}$, $\q_r \equiv \q_r^d$, and $\ddot{\q}_m \equiv \ddot{\q}_m^{ref}$ can be used in the internal dynamics which represent the dynamics of $\alpha$ and $\beta$ (roll and pitch angles of the platform):
\begin{align}
\M_{\phi \phi}  \ddot{\bphi}
 + \M_{\phi m} \ddot{\q}_m^{ref} + 
 \C_{\phi \phi} \dot{\bphi}
+
\C_{\phi m} \dot{\q}_m + \g_\phi = \bzero.
\label{eq:mot_example_internal_dyn}
\end{align}
From this equation, note that we can access the platform roll and pitch dynamics by means of $\ddot{\q}_m^{ref}$.  Assuming that $\M_{\phi m}$ is invertible\footnote{As a matter of fact, this is true, but will not be discussed in detail since this fact will not be used in the main part (Sec. \ref{sec:control}) of the paper.}, let us define $\ddot{\q}_m^{ref}$ as follows to achieve balancing of the platform.
\begin{align}
\label{eq:ddot_q_m_ref_mot_ex}
\ddot{\q}_m^{ref} = \M_{\phi m}^{-1} ( \D_\phi \dot{\bphi} + \K_\phi \bphi - \C_{\phi m}\dot{\q}_m  - \g_\phi).
\end{align}
Then, the exponential convergence of $\bphi$ is trivial, but the behavior of $\q_m$ becomes unclear. Applying Theorem \ref{thm:cascade} once again, the closed-loop dynamics of $\q_m$ is
\begin{align}
\M_{\phi m}\ddot{\q}_m + \C_{\phi m} \dot{\q}_m + \g_\phi(\q_p=\q_p^{des}, \q_m, \q_r=\q_r^{des}) = \bzero.
\label{eq:motivating_ex_mass_dyn}
\end{align}
Therefore, the moving masses are likely to experience pendulum-like behavior due to the gravity force.\footnote{To be precise, this argument is not correct as $\M_{\phi m}$ is not always positive definite. This fact may make the actual analysis complicated.}

\begin{figure}
	\centering
	\subfigure[]{
		\centering
		\includegraphics[scale=0.35]{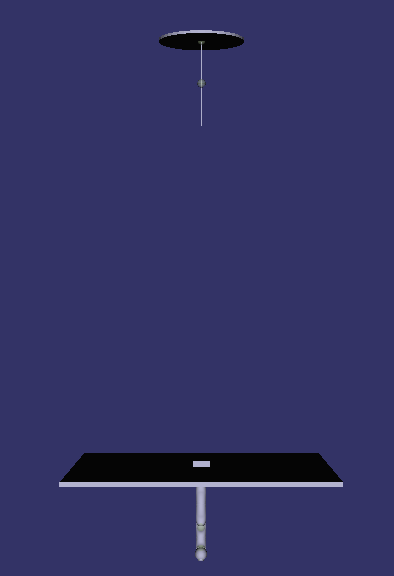}	} 
	\subfigure[]{
		\centering	
		\includegraphics[scale=0.35]{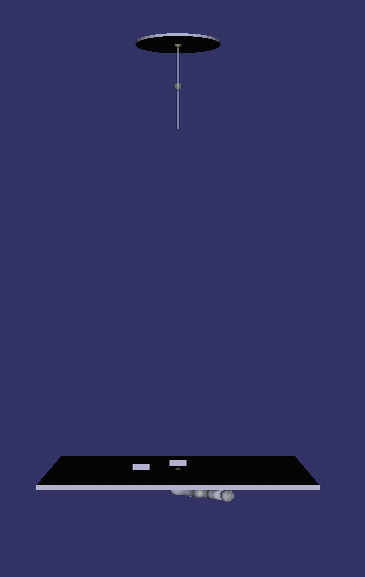}	}
	\caption{Simulation environment. (a) Initial configuration of the system. (b) Final configuration to achieve; the platform is balanced (zero roll and pitch angles); the robotic manipulator is in desired position; the moving masses are properly located to balance the static gravity of manipulator.}
	\label{fig:simul_env}
\end{figure}

\begin{figure}
	\centering
	\includegraphics[scale=0.47]{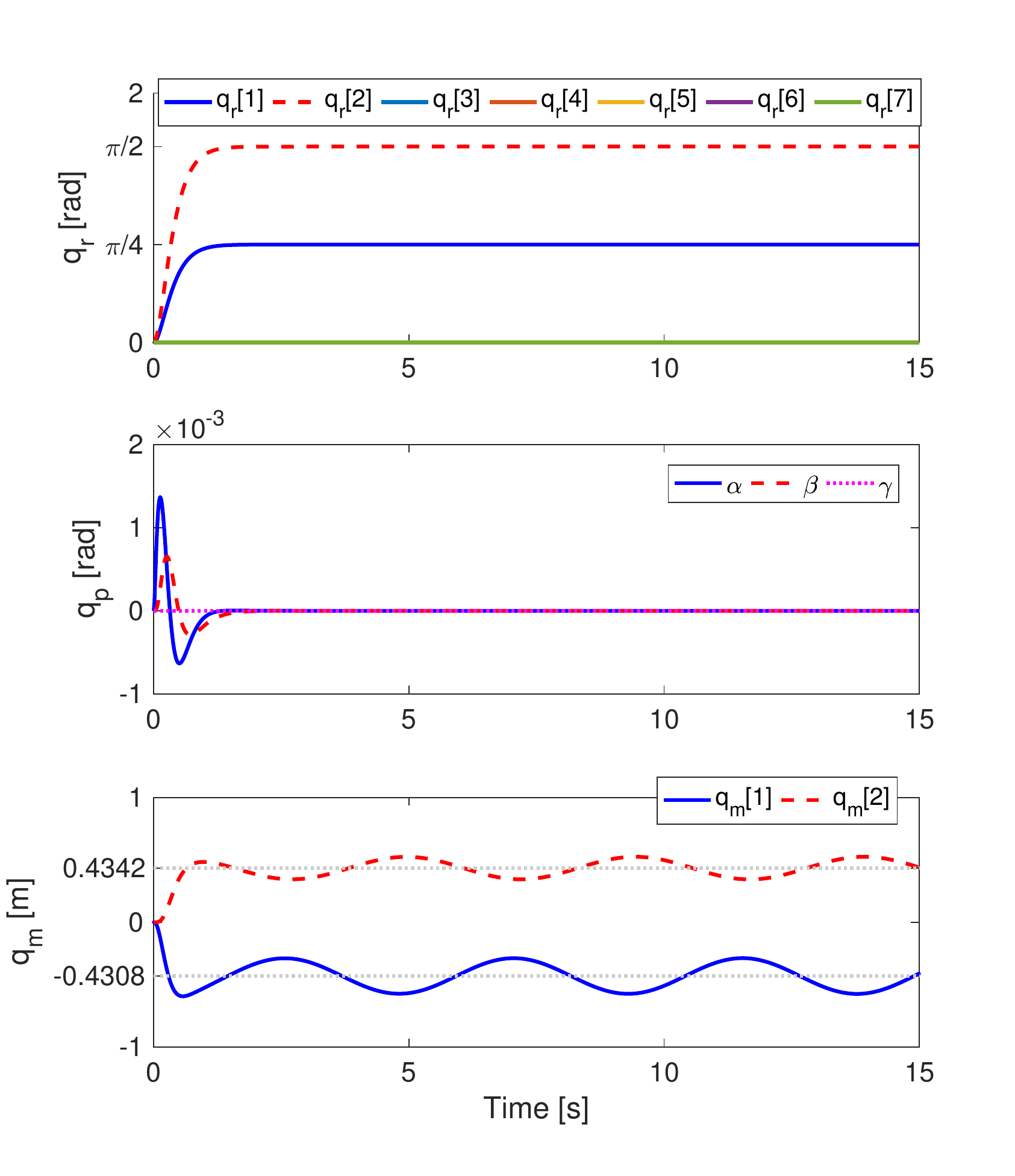}
	\caption{Simulation result for motivating example. Although the platform was balanced, the moving masses fell into an oscillatory motion.}
		\label{fig:simul_motivating_ex}
\end{figure}

The presented control approach was simulated. Fig. \ref{fig:simul_env} shows our simulation environment which is 3D extension of Fig. \ref{fig:configuration}b. Platform was hanged on a fixed base by means of wires, and a 7-DoF robotic manipulator was mounted on the platform. Two moving masses moves in $x-$, $y-$directions on the platform. The height from the platform to the base was 10 $\mathrm{m}$. The weight of each moving mass was 10 $\mathrm{kg}$, and that of platform was also 10 $\mathrm{kg}$. Total weight of the robotic manipulator was 15 $\mathrm{kg}$. Wires were massless and no friction was considered. The following scenario was simulated:
\begin{itemize}
	\item $\gamma^{des}=0$.
	\item $\q_r^{des}=[\pi/4 \; \pi/2 \; \bzero_{1 \times 5}]^T$ with zero initial condition.
	\item $\q_m^*=[-0.430813 \; 0.434241]$ is a position to balance the static gravity torque when $\q_r=\q_r^{des}$.
	\item See Fig. \ref{fig:simul_env}a (initial configuration) and Fig. \ref{fig:simul_env}b (desired configuration to achieve).
\end{itemize}
Fig. \ref{fig:simul_motivating_ex} shows the simulation results. As expected, the oscillation damping of the platform was achieved successfully. However, the moving masses converged to a oscillatory motion, which indicates the failure of the fourth requirement of the control goal. In the following section, to achieve the control goals, $\ddot{\q}_m^{ref}$ will be designed to stabilize not only the output dynamics, but also the internal dynamics.

\section{Oscillation Damping Control with Stabilization of Internal Dynamics}
\label{sec:control}


The limitation of previous approach is that the internal dynamics and output dynamics cannot be asymptotically stabilized at the same time, which is not appealing because the moving mass converges to a certain limit cycle rather than staying in the equilibrium point $\q_m^*$. Therefore, this section proposes an approach that can asymptotically stabilize the internal dynamics as well as the output dynamics.

More specifically, we seek $\ddot{\q}_m^{ref}$ that stabilizes
\begin{align}
\label{eq:closed_original_phi_dyn}
\M_{\phi \phi}\ddot{\bphi}+\M_{\phi m}\ddot{\q}_m^{ref}+\C_{\phi \phi}\dot{\q}_p+\C_{\phi m}\dot{\q}_m+\g_\phi= \bzero,& \\
\label{eq:closed_original_q_m_dyn}
\ddot{\q}_m=\ddot{\q}_m^{ref}&.
\end{align}
However, as can be deduced from the previous example, it might be hard to find such $\ddot{\q}_m^{ref}$ mainly because of the gravity $\g_\phi$. Even if $\bphi$ converges to zero in (\ref{eq:closed_original_phi_dyn}), $\g_\phi$ is not zero because of nonzero $\q_m$. Unfortunately, nonzero gravity will somehow appear in (\ref{eq:closed_original_q_m_dyn}) (recall (\ref{eq:motivating_ex_mass_dyn})), and excites the moving masses. Converse is also true. Even if we design $\ddot{\q}_m^{ref}$ so that $\q_m$ converges to a proper equilibrium in (\ref{eq:closed_original_q_m_dyn}), the gravity $\g_\phi$ is not zero due to nonzero $\bphi$. Again, nonzero gravity will excite $\bphi$ in the dynamics (\ref{eq:closed_original_phi_dyn}).

This difficulty can be overcome by expressing the dynamics (\ref{eq:whole1}) using $x,y$ components of CoM of the overall system (represented in the global frame) $\x_{c}$ instead of $\bphi$. This approach is appealing because the gravity force acting on $\x_{c}$ is zero when $\x_{c}=\bzero$. To this end, consider the following coordinate transformation:
\begin{align}
\dot{\bq} = \bT \dot{\q},
\end{align}
where 
\begin{align}
\dot{\bq} = 
\left(
\begin{array}{c}
\dot{\x}_{c} \\
\dot{\gamma} \\
\dot{\q}_m \\
\dot{\q}_r
\end{array}
\right)
\end{align}
and $\bT$ is the properly defined transformation matrix. Using this transformation, (\ref{eq:whole1}) can be expressed as
\begin{align}
\bM\ddot{\bq}+\bC\dot{\bq}+\bg=\bT^{-T} \T = \bT^{-T} \B \uu.
\label{eq:whole2}
\end{align}
Here, noting that 
\begin{align}
\bT^{-T}=
\left[
\begin{array}{cc}
* & \bzero \\
* & \bm{I}
\end{array}
\right],
\end{align}
the control input $\uu$ does not affect the CoM dynamics.

Let us now apply PFL to the output defined in (\ref{eq:y_PFL}). From 
\begin{align}
\nonumber \ddot{\y} =& \B^{T}\ddot{q} = \B^{T} \bT^{-1} (\ddot{\bq}-\dot{\bT}\dot{\q}) \\
  =& \B^{T} \bT^{-1} \left(  \bM^{-1}( -\bC \dot{\bq} - \bg  + \bT^{-T}\B \uu) - \dot{\bT}\dot{\q} \right),
\end{align}
the control input $\uu$ is defined by
\begin{align}
\nonumber \uu =& (\B^{T} \bT^{-1} \bM^{-1} \bT^{-T} \B)^{-1}  \times \\
     &\left( \B^T \bT^{-1} \left( \bM^{-1}  (\bC \dot{\bq} + \bg) + \dot{\bT}\dot{\q} \right) +  \ddot{\y}^{ref} \right).
\end{align}
Furthermore, using (\ref{eq:PFL_gamma_ref})-(\ref{eq:PFL_qr_ref}), the closed-loop dynamics after convergence of $\gamma$ and $\q_r$ is
\begin{align}
\label{eq:closed_new_com_dyn}
\bM_{cc}\ddot{\x}_{c}+\bM_{c m}\ddot{\q}_m^{ref}+\bC_{cc}\dot{\x}_{c}+\bC_{c m}\dot{\q}_m+\bg_{c}=& \bzero \\
\label{eq:closed_new_q_m_dyn}
\quad \ddot{\q}_m=&\ddot{\q}_m^{ref}
\end{align}
with $\gamma \equiv \gamma^{des}$ and $\q_r \equiv \q_r^{des}$. Similar to previous, $\bM_{cc}$ denotes the first 2 by 2 submatrix of $\bM$ (the subscript `c' stands for CoM). $\bM_{c m}$, $\bC_{cc}$, $\bC_{c m}$, and $\bg_{c}$ are defined in the same manner.

The following theorem states the main result.
\begin{thm}
Define
\begin{align}
\nonumber\ddot{\q}_m^{ref} = \D\Big(&\bM_{cm}^T (\D_c \dot{\x}_{c} + \K_c \x_c) \\
\label{eq:qm_ref_final}
       & \;\; \;- \D_m \dot{\q}_m - \K_m (\q_m - \q_m^*) \Big).
\end{align}
If the control gains $\D_{(\cdot)}$ and $\K_{(\cdot)}$ are chosen sufficiently large with $\D_c, \K_c  \gg \D_m, \K_m$, then the closed-loop dynamics is exponentially stable to $\x_{c}=\bzero$, $\q_m=\q_m^*$ which implies oscillation damping because $\bphi=\bzero$. 
\end{thm}
{\bf{Sketch of proof.}} The proof consists of two parts: The first part shows boundedness, and the second part shows exponential stability based on the first part. Before going into the proof, we keep in mind that the state $\q_m=\q_m^*$ and $\x_{c}=\bzero$ with zero derivatives is the unique equilibrium point, because $\bg_\phi=\bzero$ implies $\x_c=\bzero$.

For the first part, apply singular perturbation approach by letting $\D=\frac{1}{\epsilon_1} \bI$ and $\K_m=\frac{1}{\epsilon_1} \bI$ with $\epsilon_1>0$. Then $\ddot{\q}_m^{ref}$ itself becomes a fast variable, and converges to $\ddot{\q}_m^{ref}= \bM_{cm}^T (\D_c \ddot{\x}_{c} + \K_c\dot{\x}_c)+\dot{\bM}_{cm}^T (\D_c \dot{\x}_{c} + \K_c \x_c) - \K_m \dot{\q}_m$ exponential fast in a new time scale $t/\epsilon_1$ when $\epsilon_1 \rightarrow 0 $. This observation implies $\bM_{cm}^T (\D_c \dot{\x}_{c} + \K_c \x_c)- \D_m \dot{\q}_m - \K_m (\q_m - \q_m^*) =\bzero$, and another fast variable $\K_m (\q_m - \q_m^*)$ converges to $\bM_{cm}^T (\D_c \dot{\x}_{c} + \K_c \x_c)$ in the new time scale.  Then, the reduced system (\ref{eq:closed_new_com_dyn}) becomes
\begin{align}
\nonumber (\bM_{cc}+\bM_{cm}\bM_{cm}^T \D_c)\ddot{\x}_c + \bM_{cm}\bM_{cm}^T \K_c \dot{\x}_c + \bg_c +  \bm{\delta} = \bzero,
\end{align}
where the other terms are collected in $\bm{\delta}$ of which linearization is zero around the equilibrium point. Therefore, local asymptotic stability of $\x_c$, $\dot{\x}_c$ can be shown easily. As a result, by the Tikhonov Theorem (\cite{khalil1996noninear}), the resulting trajectory is bounded around the unique equilibrium point when $\epsilon_1$ is sufficiently small. 

For the second part, to apply singular perturbation theory in a different way, let $\D=\bI$, $\D_c=\frac{1}{\epsilon_2}\bI$, $\K_c=\frac{1}{\epsilon_2}\bI$ with $\epsilon_1 \gg \epsilon_2$ to avoid conflict with the first part. Then, (\ref{eq:closed_new_q_m_dyn}) becomes  fast dynamics with the fast variable $\bm{z}=\frac{1}{\epsilon^2}\x_c$. Since $\g_c$ vanishes with $\x_c=0$, the fast variable converges to $\bm{z}=(\bM_{cm}\bM_{cm}^T)^{-1}\big(\bC_{cm}\dot{\q} +\bM_{cm} (\D_m \dot{\q}_m + \K_m (\q_m - \q_m^*)) \big)$ and $\dot{\bm{z}}=\bzero$. Therefore (\ref{eq:closed_new_com_dyn}) becomes $\bM_{cm}\ddot{\q}_m+\bC_{cm}\dot{\q}_m =\bzero$ (note that $\bC_{mc}\dot{\x}_c=\bzero$) which is slow dynamics. Therefore, $\q_m$ will either constantly drift away or stay in a certain equilibrium point. Since the boundedness is already guaranteed from the first part, $\q_m$ can only stay in the unique equilibrium point $\q_m=\q_m^*$, which can be interpreted as exponential stability. Hence, the original system (\ref{eq:closed_new_com_dyn})-(\ref{eq:closed_new_q_m_dyn}) is exponentially stable with sufficiently small $\epsilon_1$, $\epsilon_2$ satisfying $\epsilon_1 \gg \epsilon_2$. \QEDA

In the following remarks, to further motivate (\ref{eq:qm_ref_final}),  we introduce a couple of variations which may seem plausible at a glance, but do not achieve control goals.

\begin{rem}
	Consider
	\begin{align}
	\label{eq:qm_ref_only_com}
	\ddot{\q}_m^{ref} = \bM_{cm}^T (\D_c \dot{\x}_{c} + \K_c \x_c),
	\end{align}
	which does not contain $\q_m$-related terms. Obviously, this choice will result in $\x_c \rightarrow \bzero$. However, $\x_c=\bzero$ does not always imply $\bphi=\bzero$ because $\x_c$ is the CoM represented in the global reference frame. This variation will be discussed in more detail with simulation results in next section.
	\label{rem:ext_1}
\end{rem}

\begin{rem}
One may want to extend (\ref{eq:ddot_q_m_ref_mot_ex}) to have PD term of $\q_m$ similar to (\ref{eq:qm_ref_final}). However, correct extension is not trivial as naive extension lead to non-unique equilibrium point. For example, consider
\begin{align}
\nonumber \ddot{\q}_m^{ref} = & \M_{\phi m}^{-1} ( \D_\phi \dot{\bphi} + \K_\phi \bphi - \C_{\phi m}\dot{\q}_m  - \g_\phi) \\
\label{eq:qddot_m_ref_variation}
& - \D_{m} \dot{\q}_m - \K_m (\q_m - \q_m^{*}).
\end{align}
Equilibrium point is given by solving
\begin{align}
\M_{\phi m}^{-1} ( \D_\phi \dot{\bphi} +\K_\phi \bphi)+\K_m (\q_m - \q_m^{*}) =& \bzero,\\
\g_\phi=&\bzero.
\end{align}
Unfortunately, there is no unique solution for this set of equations. This variation will be discussed in more detail with simulation results in next section.
	\label{rem:ext_2}
\end{rem}


\section{Simulation}
\label{sec:simulation}

In this section, after showing the simulation results for the variations introduced in Remarks \ref{rem:ext_1} and \ref{rem:ext_2}, the results for the proposed approach is shown.


\begin{figure}
	\centering
	\includegraphics[scale=0.47]{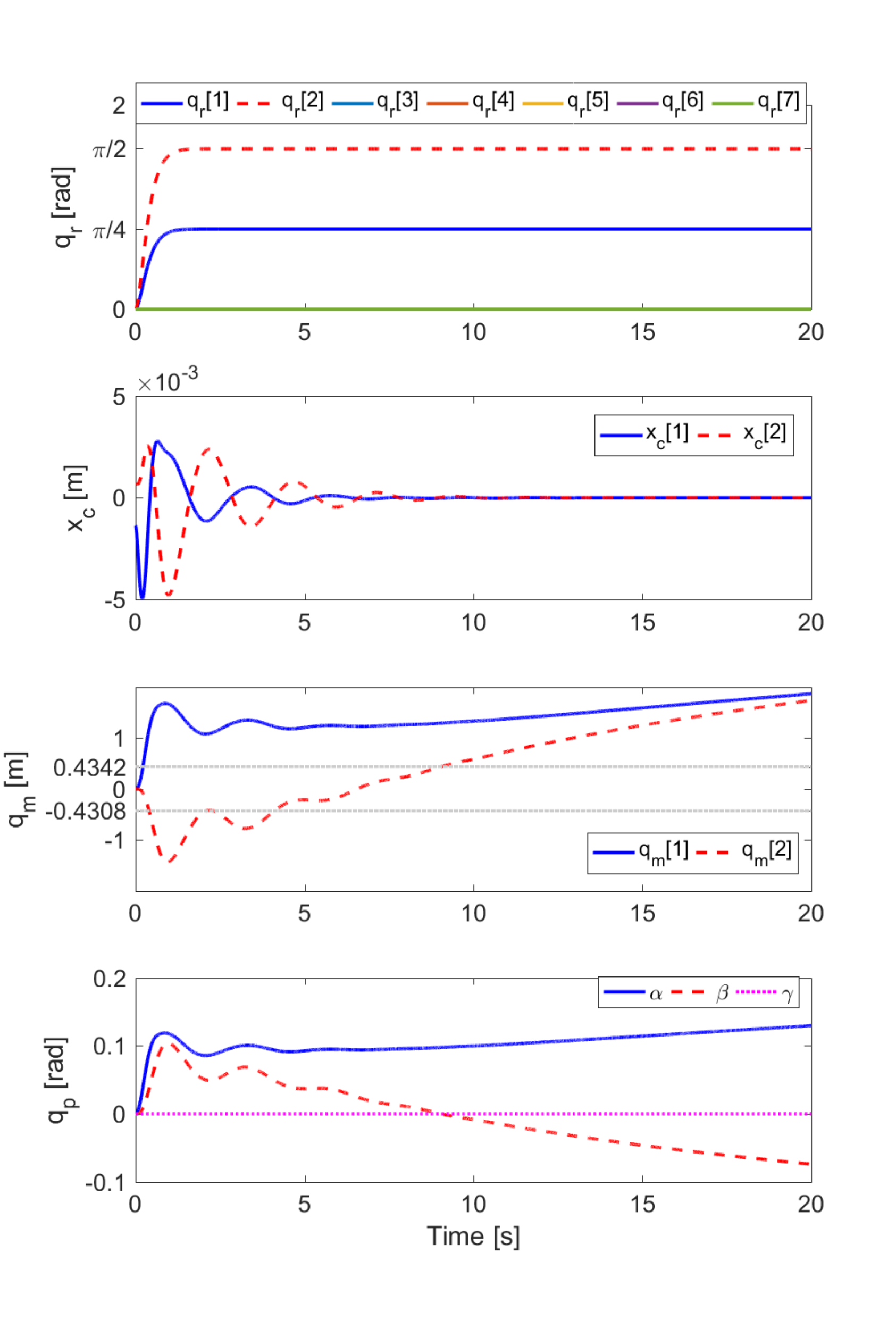}
	\caption{Simulation result with variation in Remark \ref{rem:ext_1}. Although $\x_c$ converged to $\bzero$, $\q_m$ and $\q_p$ diverged out.}
	\label{fig:variation_com}
\end{figure}

\begin{figure}
	\centering
	\includegraphics[scale=0.47]{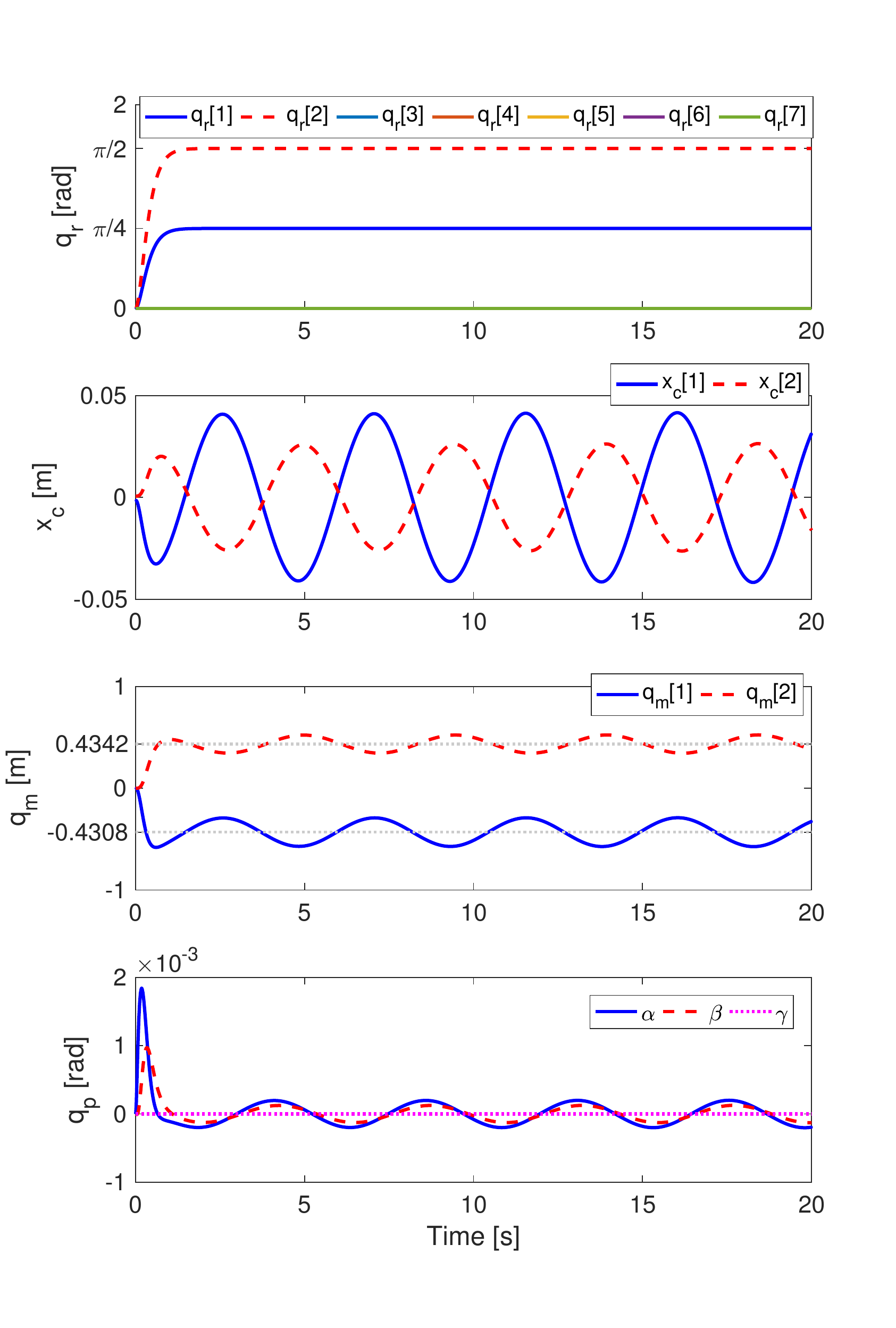}
	\caption{Simulation result with variation in Remark \ref{rem:ext_2}. Although $\q_m$ is included in $\ddot{\q}_m^{ref}$ design (\ref{eq:qddot_m_ref_variation}), $\q_m$ and $\q_p$ did not converge to the desired equilibrium point.}
	\label{fig:variation_from_orig}
\end{figure}

\subsection{Simulation result for the variation in Remark \ref{rem:ext_1}} 

In Remark \ref{rem:ext_1}, only $\x_c$ is considered in $\ddot{\q}_m^{ref}$ design. By doing so, $\x_c$ converged to zero, as shown in the second row of Fig. \ref{fig:variation_com}. However, as pointed out in Remark \ref{rem:ext_1}, $\x_c=\bzero$ does not imply $\bphi=\bzero$ (and also $\q_m=\q_m^*$). As shown in the third and fourth row of Fig. \ref{fig:variation_com}, $\q_p$ and $\q_m$ can diverge out while maintaining $\x_c=\bzero$. Note that $\q_r=\q_r^{des}$ and $\gamma=\gamma^{des}$ were still achieved because feedback linearization with pole-placement (\ref{eq:PFL_gamma_ref})-(\ref{eq:PFL_qr_ref}) was applied to these variables.

\subsection{Simulation result for the variation in Remark \ref{rem:ext_2}} 

In Remark \ref{rem:ext_2}, both $\q_\phi$ and $\q_m$ were considered in $\ddot{\q}_m^{ref}$ design. This design may seem plausible as it tries to stabilize $\q_\phi$ as well as $\q_m$ at the same time. However, as pointed out in  Remark \ref{rem:ext_2}, equilibrium point is not unique. As shown in Fig. \ref{fig:variation_from_orig}, $\q_m$ and $\q_\phi$ oscillated around the desired point, but could not converge to it.

\subsection{Simulation with the proposed approach}
\begin{figure}
	\label{fig:main_res}
	\centering
	\includegraphics[scale=0.47]{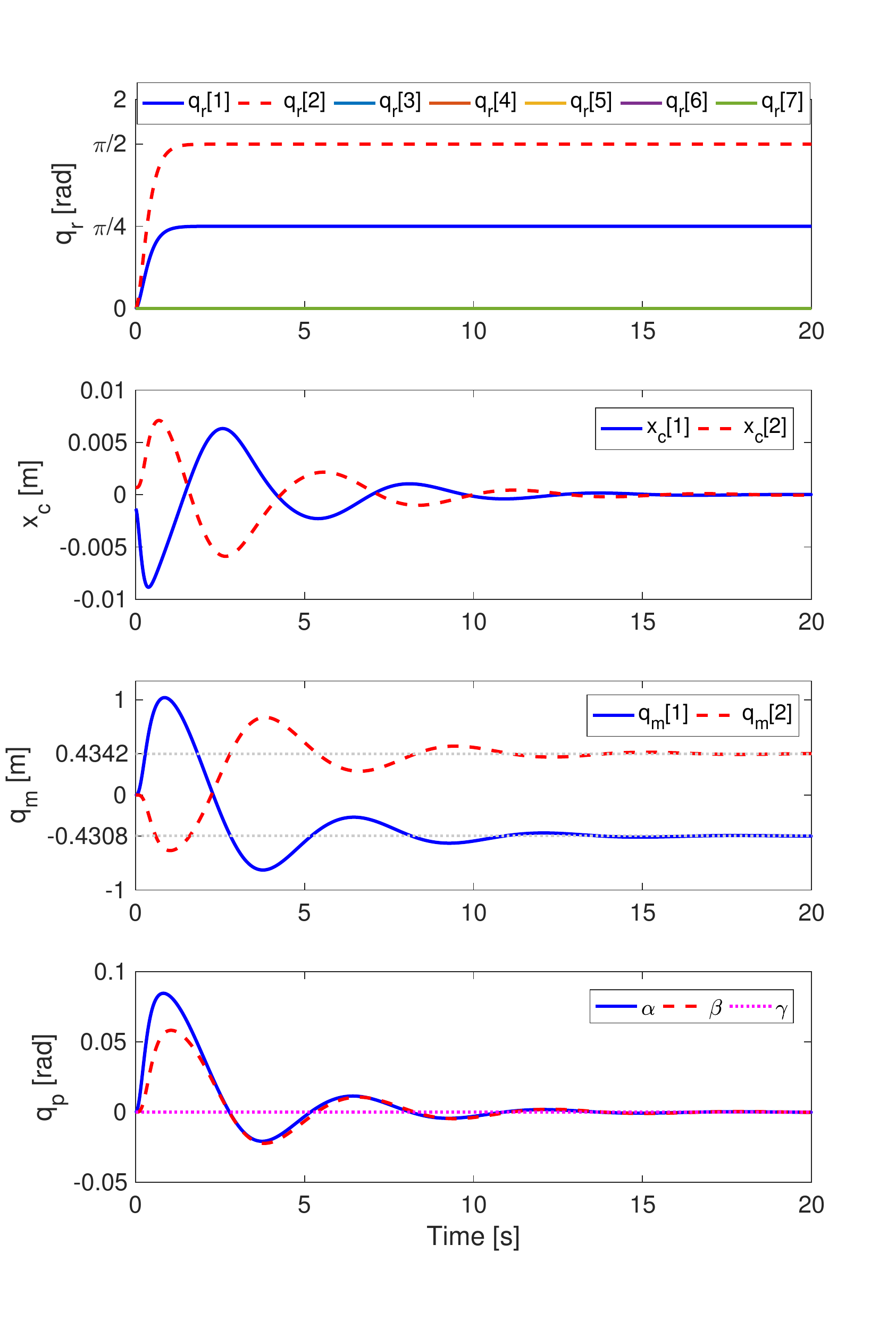}
	\caption{Simulation result with proposed approach. Both $\x_c$ and $\q_m$ converged to the desired equilibrium point. Note that this result indicates successful oscillation damping of the platform as shown in the fourth row.}
\end{figure}

Fig. \ref{fig:main_res} shows simulation results with the proposed $\ddot{\q}_m^{ref}$ in (\ref{eq:qm_ref_final}). As expected, $\x_c=\bzero$ and $\q_m=\q_m^*$ could be achieved. Note that $\x_c=\bzero$ and $\q_m=\q_m^*$ imply $\bphi=\bzero$, as shown in the fourth row of Fig. \ref{fig:main_res}. As discussed previously, $\q_r$ and $\gamma$ converged and stayed in the desired equilibrium because feedback linearzation with pole-placement technique was applied to these variables.

\section{Conclusion}
\label{sec:conclusion}

In this paper, a pendulum-like hanging manipulation platform is presented to overcome limitation of the aerial manipulation system. The main challenge in the control point of view is the oscillation damping of the platform. Since there is no actuation for the platform orientation (i.e., under-actuated), it can be controlled only indirectly. To alleviate this problem, moving masses were installed on top of the hanging platform. This paper tries to damp out the platform oscillation by defining the reference acceleration of the moving masses properly. In particular, it is shown that the extension of PFL can be used to asymptotically stabilize not only output dynamics (robotic manipulator and yaw angle of the platform) but also the internal dynamics (platform oscillation and moving mass).  The proposed approach was validated by simulation studies.

\bibliography{mybib}             
                                                   







\end{document}